\documentclass{article} 
\usepackage{iclr2022_conference,times}

\usepackage{hyperref}
\usepackage{url}
\usepackage{multirow}
\usepackage{times}
\usepackage{epsfig}
\usepackage{graphicx}
\usepackage{amsmath}
\usepackage{amssymb}
\usepackage{algorithm}
\usepackage{algorithmicx}
\usepackage{algpseudocode}
\usepackage{booktabs}
\usepackage{bm}
\usepackage{caption}
\usepackage{enumitem}
\usepackage{subcaption}
\usepackage{mathrsfs}
\usepackage{wrapfig}

\newcommand{\egno}{\textit{e}.\textit{g}.} 

\newcommand{\ieno}{\textit{i}.\textit{e}.} 

\title{Local Patch AutoAugment with Multi-Agent Collaboration}

\vspace{-1.0em}
\author{Shiqi Lin\thanks{Equal Contribution.}~, ~Tao Yu$^*$, ~Ruoyu Feng, ~Xin Li, ~Xin Jin \& Zhibo Chen  \\
University of Science and Technology of China\\
\texttt\{linsq047,yutao666,ustcfry,lixin666,jinxustc\}@mail.ustc.edu.cn, chenzhibo@ustc.edu.cn}
\begin{document}

\maketitle

\vspace{-1.5em}
\begin{abstract}
Data augmentation (DA) plays a critical role in improving the generalization of deep learning models. 
Recent works on automatically searching for DA policies from data have achieved great success.
However, existing automated DA methods generally perform the search at the image level, which limits the exploration of diversity in local regions.
In this paper, we propose a more fine-grained automated DA approach, dubbed Patch AutoAugment, to divide an image into a grid of patches and search for the joint optimal augmentation policies for the patches. 
We formulate it as a multi-agent reinforcement learning (MARL) problem, where each agent learns an augmentation policy for each patch based on its content together with the semantics of the whole image. The agents cooperate with each other to achieve the optimal augmentation effect of the entire image by sharing a team reward. We show the effectiveness of our method on multiple benchmark datasets of image classification and fine-grained image recognition (\egno, CIFAR-10, CIFAR-100, ImageNet, CUB-200-2011, Stanford Cars and FGVC-Aircraft). Extensive experiments demonstrate that our method outperforms the state-of-the-art DA methods while requiring fewer computational resources. The code is available at \href{https://github.com/LinShiqi047/PatchAutoAugment}{https://github.com/LinShiqi047/PatchAutoAugment}.

\end{abstract}

\section{Introduction}

\begin{wrapfigure}{r}{0.40\textwidth}
  \vspace{-20pt}
  \begin{center}
  \includegraphics[width=0.40\textwidth]{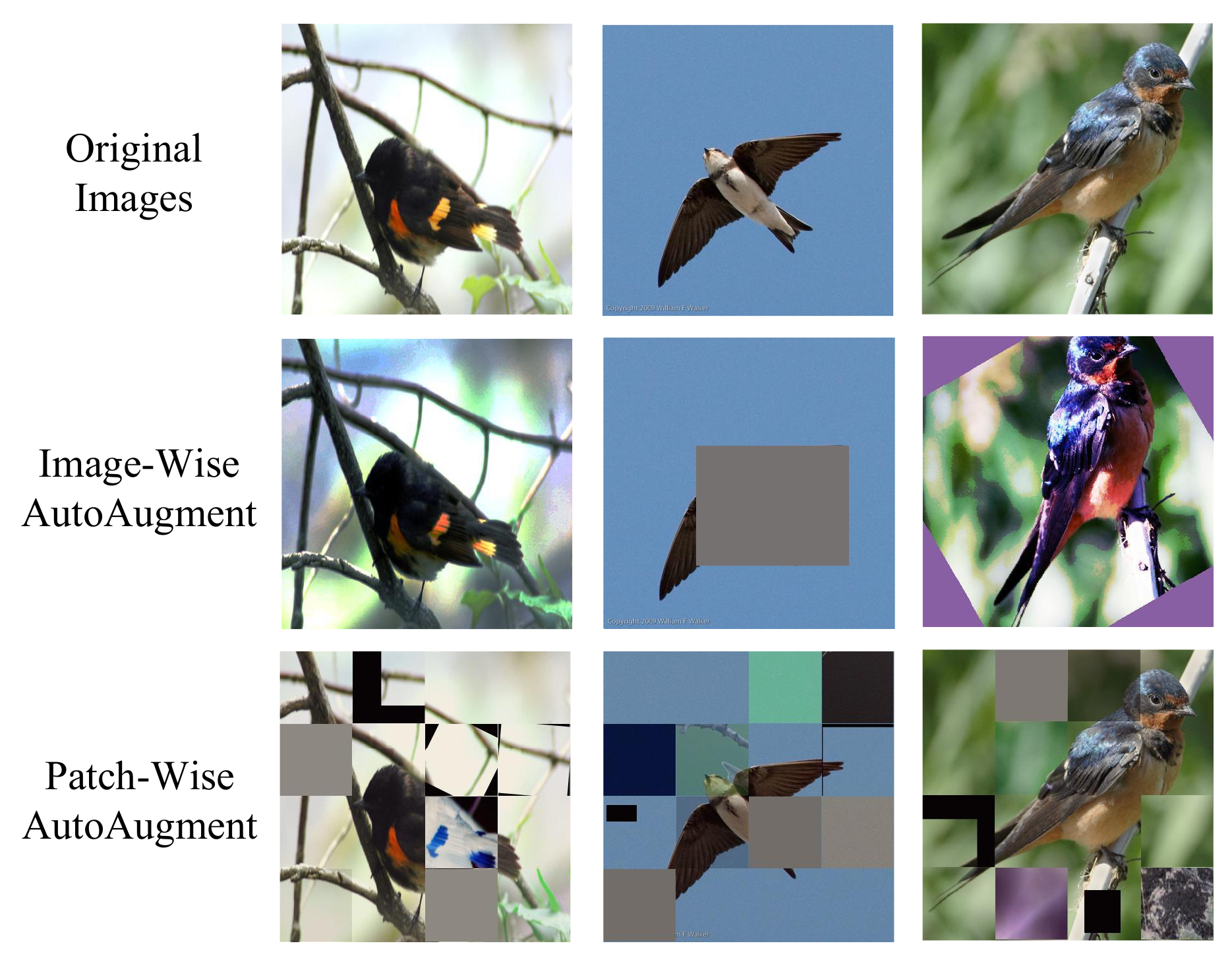}
  \end{center}
  \vspace{-10pt}

  \caption{Illustration of different automated augmentation policies. We show the examples  processed by image-wise automated DA, AutoAugment (middle row) and processed by our proposed Patch AutoAugment (bottom row). }
  \vspace{-10pt}

  \label{fig:AA_PAA}
  
\end{wrapfigure}

Data Augmentation (DA) has been widely used to alleviate the overfitting risk in training deep neural networks by appropriately enriching the diversity of training data \citep{shorten2019survey,gontijo2020affinity}.
Notable DA methods improve the performance and robustness of the neural networks, such as rotation, Mixup \citep{zhang2017mixup} and Cutmix \citep{yun2019cutmix}. 
But these approaches are typically handcraft and require human prior knowledge, which causes weak transferability of DA across different datasets.
To relieve the dependence on manual design and further explore more adaptive augmentation, AutoAugment (AA) \citep{cubuk2018autoaugment}, as a new DA paradigm, is proposed to automate the search of the optimal DA policies (\ieno, DA operation, probability and magnitude) from the training dataset. To be specific, AA trains a proxy model with the augmentation policy generated by a controller, which is updated through reinforcement learning using validation accuracy as the reward signal.
In spite of the superior performance of AA, its optimization procedure is computationally intensive due to the need to evaluate thousands of policies. Therefore, to reduce the search costs, multiple automated DA approaches \citep{lim2019fast,ho2019population,hataya2019faster,zhang2019adversarial,lin2019online} have been proposed. For example, \citep{lim2019fast} employs density matching as a search method to accelerate the policy search, and \citep{zhang2019adversarial} introduces adversarial learning to organize the target network training and augmentation policies search in an online manner.

Yet the aforementioned automated DA methods all search for policies at the image level. They ignore the exploration of diversity in local regions, which may result in insufficient diversity of dataset and limit the benefits of DA \citep{gontijo2020affinity}.
In addition, due to this coarse-grained augmentation, they may damage key semantic information and introduce ambiguity into the training process (see a simple example in Figure \ref{fig:AA_PAA} row 2, column 2).
With those in mind, it is necessary to automatically search for the optimal augmentation policies for different regions by taking regional diversity into account.
One straightforward idea is to directly apply image-wise automated DA methods in different regions. But such an intuitive solution ignores the contextual relationship between regions, which may lead to the non-globally optimal effectiveness of DA policies across the entire image. Besides, it may encounter an extremely high computational cost due to the need of optimizing multiple policies for regions respectively.

In this paper, we propose a new approach, named Patch AutoAugment (PAA) (see the last row of Figure \ref{fig:AA_PAA}), to address the above-mentioned problems.
We first divide an image into a grid of patches to increase the flexibility of representations for different regions and take “patch” as the basic control unit. Then, we model the search for the augmentation policies of patches as a fully cooperative multi-agent task, and we leverage a multi-agent reinforcement learning (MARL) algorithm where agents cooperatively learn the policies. 
Specifically, based on the content of each patch and the semantics of the entire image, the agent searches for a policy in terms of choosing which transformation to apply out of the pre-defined DA operations. To encourage our policy networks to adaptively learn the beneficial augmentation policies for the target network, we use the feedback of the target network as the team reward signal to guide the policy networks to learn on the fly. All agents wind up benefiting from two mechanisms (\ieno, parameter sharing and centralized training with decentralized execution) in MARL. In this way, all agents collaboratively and 
parallelly learn policies to further achieve the joint optimal DA policy across the whole image and alleviate the computational cost. 

In summary, our contributions can be summarized as follows: 1) We pinpoint that exploring diversity in local regions is important for automated learned DA approaches. To our best knowledge, we are the first to propose a more fine-grained automated DA approach, that searches for the optimal policies for the patches according to the content of the patch and the semantics of the entire image.
2) To further achieve the joint optimal policy across the image, we model the DA policy search of patches as a fully cooperative multi-agent task, and adopt a multi-agent reinforcement learning algorithm for Patch AutoAugment by considering the contextual relationship between the patches.
3) Our visualization results provide some insights to the DA community on which augmentation operation should be chosen for patches with different content during the whole training process.

\begin{figure*}[t]
\vspace{-2.0em}
	\setlength{\abovecaptionskip}{0pt} 
\setlength{\belowcaptionskip}{-0pt}
	\begin{center}
		\includegraphics[width=1\linewidth]{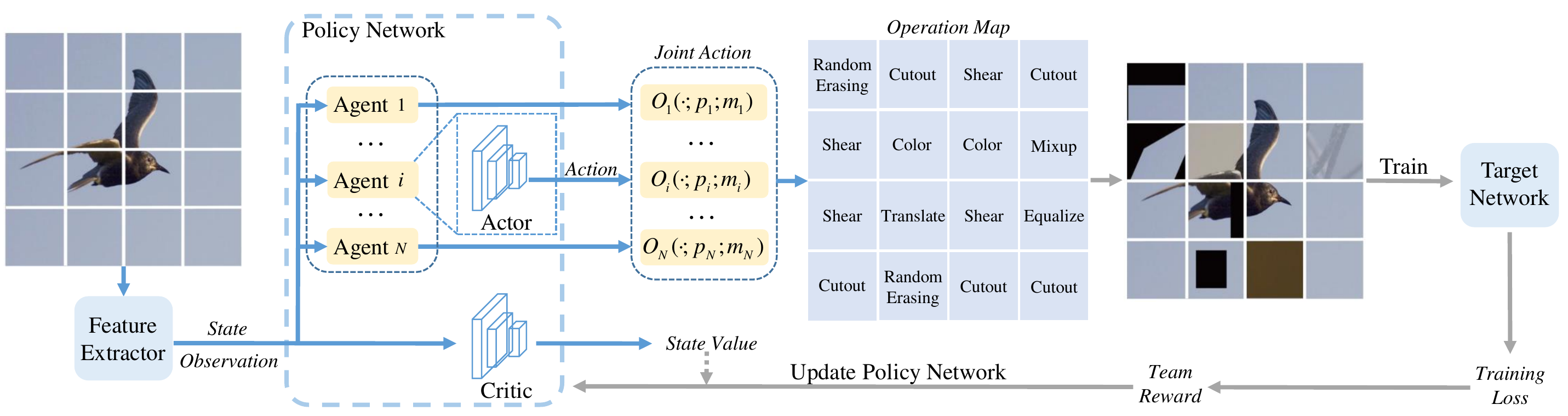}   
	\end{center}

	\caption{The framework of Patch AutoAugment (PAA). We divide an image into a grid of patches and assign an agent to each patch for selecting the optimal augmentation operation according to the patch content together with the whole image semantics. The agents cooperate with each other to achieve the joint optimal DA results by sharing a team reward through multi-agent reinforcement learning (MARL). Specifically, each agent outputs $O_i(\cdot;p_i;m_i)$ operation performed on patch $i$, which includes two parameters: probability of calling the operation $p_i$ and the magnitude $m_i$. Note that PAA is co-trained with the target network.}
	\label{fig:PAA}
\vspace{-0.6    em}
\end{figure*}


\section{Related Work}
\subsection{Data Augmentation}

Despite the remarkable performance of deep learning models in computer vision tasks, they often suffer from overfitting. Data augmentation (DA) as an effective technique has been proved to improve the generalization ability of deep learning models \citep{shorten2019survey}. Previous works \citep{cubuk2020randaugment,gontijo2020affinity} indicate that the main benefit of DA arises from increasing the diversity of images.
Popular techniques, such as rotation, flipping, color transformation, have been performed as commonly used augmentation methods. Recently, thanks to the skillful design of human experts, many DA methods (\egno, Cutout \citep{devries2017improved}, Mixup \citep{zhang2017mixup} and CutMix \citep{yun2019cutmix}) have been proposed and show significant performance. However, these manually designed methods require additional human prior knowledge on the dataset and sometimes they are limited to certain datasets and target tasks. 
Naturally, automatically finding DA methods from data have emerged to overcome the limitations of dependence on manual cumbersome exploration. Some works use generative adversarial networks (GANs) to directly generate training data \citep{shrivastava2017learning,tran2017bayesian}. 

Furthermore, recent studies aim to automate the search for augmentation policies that choose the optimal transformations for training images. AutoAugment (AA) \citep{cubuk2018autoaugment} adopts a controller to generate an augmentation policy which is used to train a proxy network, then gets the validation accuracy as the reward signal to update the controller using reinforcement learning. Unfortunately, the evaluation of thousands of policies makes AA computationally expensive. Therefore, multiple automated DA approaches focus on reducing the huge complexity and have achieved great progress. For example, PBA \citep{ho2019population} employs hyperparameter optimization, Fast AA \citep{lim2019fast} uses a density matching algorithm and Adversarial AA \citep{zhang2019adversarial} proposes an adversarial framework to jointly optimize the target network and the augmentation network. However, these automated augmentation methods perform the search at the image level, \ieno, they use the same policy on the whole image. It inevitably ignores the diversity of different regions in an image, which limits the diversity of data increased by data augmentation, and sometimes causes the damage of critical semantic information. In contrast, our method takes diversity in local regions and contextual relationship into account, aiming to search for augmentation policies for multiple regions and achieve the joint optimal DA effect across the entire image through MARL.

Our proposed method is conceptually orthogonal to most region-based DA methods where DA transformations are performed in a non-automated way. For example, RandomErasing \citep{zhong2020random}, CutMix \citep{yun2019cutmix} perform cropping or replacement operation on a randomly selected rectangle region. Some works further exploit class activation map \citep{zhou2016learning} or saliency map \citep{zhou2015salient} to select representative regions which are augmented by a randomly selected DA operation (\egno, SaliencyMix \citep{uddin2020saliencymix}, SnapMix \citep{huang2020snapmix}) and KeepAugment \citep{gong2020keepaugment}). Yet our proposed method automatically searches for the augmentation transformations based on the given datasets and tasks.

\vspace{-0.5em}
\subsection{Multi-Agent Reinforcement Learning}

The most significant characteristic of MARL is the cooperation between agents \citep{tampuu2017multiagent,ma2008combining,foerster2017learning} which is distinct from directly applying reinforcement learning (RL) algorithm to multi-agent systems. Due to the limited observation and action of a single agent, cooperation is necessary in the reinforced multi-agent system to achieve the common goal. Compared with independent agents, cooperative agents can improve the efficiency and robustness of the model \citep{neto2005single,zhang2019multi,busoniu2008comprehensive}. Many vision tasks use MARL to interact with the public environment to make decisions, with the goal of maximizing the expected total return of all agents, such as image segmentation \citep{liao2020iteratively,han2019grid,ma2020boundary}, image processing \citep{furuta2019pixelrl}.

\vspace{-0.5em}
\section{Proposed Method}
\label{section:method}

As above mentioned, we aim to search augmentation policies for the patches to explore more augmentation diversity. Due to the need to consider the effectiveness of DA, it is of great necessity to take the patch content and the contextual relationship between patches into account. Therefore, we model the search problem as a multi-agent Markov decision process and use multi-agent reinforcement learning (MARL) algorithm. Specifically, the search for patch policies is based on the content of the patch together with the semantics of the image, and policies are encouraged to coordinate to achieve the joint optimal DA policy across the whole image. In this section, we first describe the preliminaries of MARL, then elaborate on our augmentation policy formulation and modelling. Furthermore, we summarize the framework of Patch AutoAugment.

\subsection{Preliminaries of Multi-Agent Reinforcement Learning}

We first introduce the preliminaries of reinforcement learning (RL). RL models the decision-making problem as a Markov decision process (MDP) which is presented with a tuple $({\mathcal{S}},{\mathcal{A}},{P},R,\gamma,T )$. In RL framework, given the state $s \in {\mathcal{S}}$, the agent takes an $a \in {\mathcal{A}}$ according to its policy ${\pi} (a\left| {s)} \right.:{\mathcal{S}}\times {\mathcal{A}} \to [0,1]$ and then receives a reward $r:{\mathcal{S}} \times {\mathcal{A}} \to \mathbb{R}$. The environment moves to the next state with a transition function denoted as ${P}:{\mathcal{S}} \times {\mathcal{A}} \times {\mathcal{S}} \to [0,1]$.
$\gamma \in (0,1]$ is a discount factor and $T$ is a time horizon. The agent aims to maximize the long-term reward $R$ over $T$ steps to learn the optimal policy ${{\pi ^*}}$.

Furthermore, multi-agent reinforcement learning (MARL) considers a group of $N$ agents, denoted as $\mathcal{N}$, operating cooperatively in a shared environment towards a common goal. It can be formulated as a multi-agent MDP (MAMDP) \citep{boutilier1996planning} represented with a tuple $({\mathcal{S}},\{{\mathcal{O}_i\}{{^{{N}}_{i=1}}}},\{{\mathcal{A}_i\}{{^{{N}}_{i=1}}}},{P},R,\gamma,T )$. Here, $\mathcal{S}$ describes the shared state space and $\{\mathcal{O}_1,\cdot\cdot\cdot,\mathcal{O}_{{N}}\}$ is a set of observations for agents. In MARL configuration, each agent receives a private observation correlated with part of state, \ieno, $o_i: \mathcal{S} \to \mathcal{O}_i$. According to the global state $s$ and its observation $o_i$, the agent takes its action ${a_i}\in{ \mathcal{A}_i}$ based on its policy ${\pi_i}{{(a_i}{|o_i,s)}}$, $\mathcal{A}_i$ is the action space for the $i$-th agent and $\bm{\mathcal{A}}=\{\mathcal{A}_1,\cdot\cdot\cdot,\mathcal{A}_N\}$ denotes the joint action space. Then, the joint action $\bm{a}={a}_1{\times}\cdot\cdot\cdot{\times}{a}_{{N}}$ produces next state according to transition function ${P}:{\mathcal{S}} \times \bm{\mathcal{A}} \times {\mathcal{S}} \to [0,1]$ and the environment gives the team reward $r:{\mathcal{S}} \times \bm{\mathcal{A}} \to \mathbb{R} $ to agents $\mathcal{N}$. The objective of each agent is to maximize the total accumulative reward $R$ to cooperatively learn the globally optimal policy $\bm{\pi ^*}=\{{\pi{_1^*}},\cdot\cdot\cdot,{\pi{^*_{{N}}}}\}$ that consists of the optimal policy of each agent.
In this paper, as mentioned before, we employ MARL to search optimal policy for each patch, and achieve the optimal effectiveness of DA across the entire image to further improve the performance of the target network as possible.

\subsection{Patch AutoAugment}

In our proposed Patch AutoAugment (PAA), we formulate the task of policy search for the patches as a cooperative multi-agent decision-making problem and adopt multi-agent reinforcement learning (MARL) to solve it. In the following, We clarify the detailed formulation $i$) the \textit{state}, \textit{observation} and \textit{action} modeling for the policy, $ii$) an effective team reward function design and $iii$) the detailed MARL algorithm for policy learning) of Patch AutoAugment. 

\textbf{Policy Modeling.} 
As illustrated in Figure \ref{fig:PAA}, given the original input batch $\bm{x}$ and the corresponding label $\bm{y}$ (\ieno, $\bm{\{x,y\}}=\{x_j,y_j\}{_{j=1}^b}$ and $b$ is the batch size), we divide an image $x_j$ into ${N}$ equal-sized and non-overlapping patches, denoted as $x_j=\{{\mathcal{P}{_j^i}\}{_{i=1}^{{N}}}}$ where ${\mathcal{P}{_j^j}}$ is the $i$-th patch of the image $x_j$. In our proposed method, we aim to search the augmentation policy for each patch. Therefore, in MARL formulation, the augmentation policy of the patch ${\mathcal{P}{^i}}$ is controlled by an agent $i \in \mathcal{N}$ and we detail the \textit{state}, \textit{observation} and \textit{action} for the augmentation policy as below.

\quad \textit{State.} As aforementioned, the selection of augmentation operation for a patch is closely bound up with the contextual relationship between regions. Therefore, the augmentation policy needs to perceive the image semantics and we take the deep features of the whole image extracted by a backbone (\egno, ImageNet pre-trained ResNet-18 \citep{he2016deep}) as the global {state} $s$ which is visible to all agents.

\quad \textit{Observation.} Apart from capturing {state} (\ieno, the global information), the agent only use their own observation (\ieno, the local information) which is invisible to other agents. The augmentation policy seeks to choose augmentation operation based on the content of the patch and the {observation} is generally part of the {state}. Considering all these factors, we utilize the deep features of the $i$-th patch as {observation} $o_i$, which are extracted by the same backbone as the state extractor.

\quad \textit{Action.} The augmentation policy is responsible for choosing which transformations to apply from pre-defined operations. Following the previous automated DA methods \citep{cubuk2018autoaugment,lim2019fast}, we define the fifteen operation functions (\ieno, ShearX/Y, TranslateX/Y, Rotate, Invert, Equalize, Solarize, Posterize, Contrast, Color, Brightness, Sharpness, RandomErasing, Cutout, Mixup, Cutmix) to construct the action space $\mathcal{A}$. Given the {state} $s$ and the {observation} $o_i$, according to policy ${\pi_i}{{(a_i}{|o_i,s)}}$, each agent $i$ determines an {action} $a_i(
\cdot)\in \mathcal{A} $, which is the operation performed on the patch. Each operation is associated with two hyperparameters: probability $p$ and magnitude $m$. In order to dramatically and effectively reduce the action space for augmentation policy, similar with  \citep{cubuk2020randaugment,ho2019population,lim2019fast}, we take a fixed probability and magnitude schedule. Among them, the probability of applying the operation is sampled from the uniform distribution (\ieno, $p_i\sim U(0,1)$. Following \citep{ho2019population}, we employ the same linear scale to be the magnitude schedule. In summary, the processed $i$-th patch in image $x_j$ is denoted as $\widetilde{P}{_j^i}=a_i({P}{_j^i})$ with the probability $p_i$ and the magnitude of $a_i(\cdot)$ is $m_i$, otherwise $\widetilde{P}{_j^i}={P}{_j^i}$.

Note that most operations are label-invariant, except Mixup \citep{zhang2017mixup} and CutMix \citep{yun2019cutmix} are label-disturbing operations that combine different patches as well as their labels. We take Mixup as an example, then $\mathcal{\widetilde{P}}{_j^i}=\lambda\mathcal{{P}}{_j^i}+(1-\lambda)\mathcal{{P}}{_t^i}$ with probability $p_i$, where $\mathcal{{P}}{_t^i}$ is a patch from another image $x_t,t\neq{j}$, $\lambda \sim Beta(\alpha,\alpha)$, for $\alpha \in (0,\infty)$, and the one-hot label is modified as $\widetilde{y}_j=\frac{\lambda}{N} y_j+\frac{(1-\lambda)}{N}y_t$. Until all patches are processed by the corresponding operations chosen by the augmentation policies, we obtain the image augmented by our PAA, denoted as $\widetilde{x}_j=\{\mathcal{\widetilde{P}}{_j^1},\cdot\cdot\cdot,\mathcal{{\widetilde{P}}}{_j^{{N}}}\}$, and the final label $\widetilde{y}_j$. More operation details are shown in Appendix \ref{appendix:operation}.

\textbf{Reward Function.} 
The reward function is of importance to guide the agents to learn so that they follow desired behaviors. The previous work, Adversarial AA (AdvAA) \citep{zhang2019adversarial}, attempts to increase the training loss of the target network to generate harder augmentation policies and explore the weakness of the target network. Inspired by AdvAA, we reformulate the reward design appropriately under our configuration. In our proposed PAA, the common objective of all agents is to improve the performance of the mainstream target task through enhancing the benefits of DA. Therefore, we compare the feedback of target network $\phi(\cdot)$ on the augmented data processed by our proposed PAA $\widetilde{\bm{x}}$ with the original data $\bm{x}$ and take their difference on the training losses as the reward for the policy in MARL, as in Eq. (\ref{eq:reward}):
\begin{equation}
\label{eq:reward}
{r} =  l(\phi (\bm{x}),\bm{y})-l(\phi (\bm{\widetilde x}),\bm{\widetilde y}),  
\end{equation}

where $\bm{x}$ and $\bm{y}$ denote raw inputs and labels in supervision tasks. In our PAA model, all agents are encouraged to cooperate to achieve the common goal, thus we adopt the team reward function design (\ieno, the shared reward mechanism in MARL) as Eq. (\ref{eq:reward}) for all agents to make the joint augmentation policy achieve the optimal effectiveness. 

\textbf{Policy Learning.} 
Here, we introduce the training for the augmentation policies mentioned above. Considering that the action space is discrete, we adopt multi-agent Advantage Actor-Critic algorithm \citep{lowe2017multi} to learn the augmentation policies and encourage the coordination behaviors.  In MARL, the framework of centralized training with decentralized execution \citep{foerster2018counterfactual,rashid2018qmix} is generally adopted. More concretely, each agent $i$ has an actor which is to learn discrete policy $\pi_i(a_i|o_i,s)$ and agents share a common critic which aims to estimate the value of global state $V^{\bm\pi}(s)$. And we use the centralized critic to train decentralised actors. Here, we reformulate it appropriately for our task. We model the centralized action-value Q function that takes the actions of all agents in addition to state information $s$ and outputs the Q-value for the team, formulated as:

\begin{equation}
\label{eq:q}
Q^{\bm{\pi}}(s,\bm{a}) = {\mathbb{E}_\pi }[{R_t}|s,a_1,\cdot\cdot\cdot,a_N],
\end{equation}

where $\bm{a}$ is the joint action of all agents $\bm{a}=\{a_i,\cdot\cdot\cdot,a_N\}$ and ${R_t} = \sum\nolimits_{l = 0}^T {{\gamma ^l}{r_{t + l}}}$ is the long-term discounted reward. Then, the advantage function on the augmentation policy is given as follows:
\begin{equation}
    A^{\bm{\pi}}(s,\bm{a})=Q^{\bm{\pi}}(s,\bm{a})-V^{\bm{\pi}}(s).
\end{equation}

And we use $\varphi$ to denote the critic network parameters. We take the square value of the advantage function $A^{\pi}$ as the loss function to update $\varphi$:
\begin{equation}
   L(\varphi) = {\left( A^{\bm\pi}(s,\bm{a}) \right)^2}.
\end{equation}
Besides, to further achieve the ability to cooperate, similar to \citep{foerster2018counterfactual,rashid2018qmix}, the parameters of the actor networks of all agents are shared, denoted as $\theta$. In addition, the loss function for updating $\theta$ is defined as:

\begin{equation}
{L}(\theta) =  - \log {\pi _{{\theta}}}(\bm{a}|s)A^{\bm{\pi}}(s,\bm{a}).
\end{equation}

\textbf{Framework Summary.}
In this part, we summarize the overall framework of our proposed PAA. As shown in Figure \ref{fig:PAA}, PAA first divides an image into a grid of patches. Then, we use a feature extractor to obtain the deep features of the whole image as the state. Each agent draws its individual observation, \ieno, the deep features of the patch. According to the global state (\ieno, the semantics of the entire image) and the local observation (\ieno, the content of the patch), the actor networks output the augmentation operations of patches to further construct the joint operation map performed on the whole image. The augmented mini-batch is processed by our proposed PAA and then we input it to the target task network for parameters updating. Moreover, the feedback of the target network is used as the team reward signal to update the policy network. 


\begin{table*}[t]
\vspace{-1.0em}
\begin{center}
\caption{Test set accuracy (\%) on CIFAR-10 and CIFAR-100. The results of our proposed PAA is the average accuracy ($\pm${standard deviation}) over four random runs.}
\label{table:accuracy on cifar}
\scalebox{0.93}{
\begin{tabular}{c|c|c c c c c c|c}
\hline
\hline
Dataset&Model & Baseline & CutOut & Mixup & CutMix & AA & FastAA & PAA \\
\hline
\multirow{5}{*}{CIFAR-10} & WRN-28-10 & 96.1 & 96.9 & 97.1 & 97.2 & 97.4 & 97.3 & \textbf{97.5}\scriptsize$\pm 0.1$  \\
&SS(26 2x32d) & 96.4 & 97.0 & 97.2&97.3 &97.5 & 97.3 &\textbf{97.6\scriptsize$\pm 0.1$}\\
&SS(26 2x96d) &97.1 & 97.4 & 97.7 & 97.8 &97.7 &97.7 &\textbf{98.1\scriptsize$\pm 0.1$}\\
&SS(26 2x112d) &97.2 & 97.4 & 98.0 & 98.0 &\textbf{98.1} &98.0 &\textbf{98.1}\scriptsize$\pm 0.1$\\
&Pyramid+SD & 97.3 &97.7 & 98.0 & 98.1 &{98.5} &98.2 & \textbf{98.6\scriptsize$\pm 0.1$} \\
\hline
\multirow{3}{*}{CIFAR-100} &WRN-28-10&81.2&81.6&82.1 &82.8 & 82.9&82.7&\textbf{83.4\scriptsize$\pm 0.3$}\\
&SS(26 2x96d) &82.9 & 84.0 &85.4 &85.6 &85.7 &85.1 &\textbf{85.9\scriptsize$\pm 0.2$}\\
&Pyramid+SD & 86.0 &87.8 & 88.5 & 88.9 &\textbf{89.3} & 88.1 & 89.2\scriptsize$\pm 0.1$ \\
\hline
\hline
\end{tabular}}
\end{center}  

\end{table*}

\begin{table}[t]
\vspace{-0.5em}
\caption{Validation set Top-1 / Top-5 accuracy (\%) on ImageNet.}
\label{table:accuracy on imagenet}
\begin{center}
\setlength{\tabcolsep}{1.6mm}{
\scalebox{0.93}{
\begin{tabular}{c|c c c c c |c}
\hline
\hline
Method & Baseline & Mixup & CutMix & AA &FastAA  & PAA \\
\hline
ResNet-50 & 76.3 / 93.1 & 77.0 / 93.4 & 77.2 / 93.5 & 77.6 / 93.8 & 77.6 / 93.7 & \textbf{78.3\scriptsize{$\pm 0.3$}} \textbf{/} \textbf{94.1}\scriptsize{$\pm 0.2$} \\
ResNet-200 & 78.5 / 94.2 & 79.6 / 94.8  & 79.9 / 94.9 & 80.1 / 95.0 & 80.6 / 95.3 & \textbf{81.0\scriptsize$\pm 0.3$} \textbf{/} \textbf{95.2}\scriptsize$\pm 0.2$ \\

\hline
\hline
\end{tabular}}}
\vspace{-1.0em}
\end{center}
\end{table}

\section{Experiments}
\label{subsection:exp}

{\subsection{Experiment Overview}}

In this section, to study the effectiveness of Patch AutoAugment (PAA), we experiment with core image classification and fine-grained image recognition tasks. Exactly, we focus on CIFAR-10, CIFAR-100 \citep{krizhevsky2009learning} and ImageNet \citep{deng2009imagenet} datasets as well as three fine-grained object recognition datasets, \ieno, CUB-200-2011 \citep{wah2011caltech}, Stanford Cars \citep{krause20133d} and FGVC-Aircraft \citep{maji2013fine}. We describe the datasets in detail in Appendix \ref{appendix:dataset}. We compare PAA with baseline pre-processing, Cutout \citep{devries2017improved}, Mixup \citep{zhang2017mixup}, Cutmix \citep{yun2019cutmix}, AutoAugment (AA) \citep{cubuk2018autoaugment} and Fast AutoAugment (FastAA) \citep{lim2019fast}. The baseline follows \citep{zoph2018learning,yamada2018shakedrop,gastaldi2017shake}: standardizing the data, horizontally flipping with 0.5 probability, zero-padding
and random cropping. More details about our policy network architectures and model hyperparameters are supplied in Appendix \ref{appendix:model} and \ref{appendix:parameters}, respectively. 
Moreover, we set the number of patches $N$ to $4$ for CIFAR tasks, and $N=16$ for ImageNet together with fine-grained recognition datasets. To ensure the reliability of our experiments, we run each experiment four times using different random seeds.

\subsection{Results and Analysis}

\textbf{Classification Results on CIFAR-10 and CIFAR-100.} 
For CIFAR-10 and CIFAR-100, we examine on Wide-ResNet-28-10 (WRN-28-10) \citep{zagoruyko2016wide}, Shake-Shake (SS) \citep{gastaldi2017shake} and Pyramid-Net+ShakeDrop (Pyramid+SD) \citep{han2017deep,yamada2018shakedrop} models. The results are reported in Table \ref{table:accuracy on cifar}, which shows the proposed approach consistently outperforms several state-of-the-art DA methods. We observe that the improvement of performance is relatively slight, due to the small image size of CIFAR which is $32\times{32}$. In the following, we further apply our proposed PAA on datasets with larger image sizes and other networks.

\textbf{Classification Results on ImageNet.} As shown in Table \ref{table:accuracy on imagenet}, we evaluate our method on ResNet-50 and ResNet-200 \citep{he2016deep} backbone on ImageNet, and our PAA significantly improves the performance of the target networks. The results further demonstrate that our proposed method is an effective DA technique for consistent and expressive benefits for datasets with larger image sizes.

\textbf{Effectiveness of Fine-grained Classification.} Furthermore, we evaluate our proposed method on fine-grained image recognition tasks. According to previous work \citep{du2020fine,chen2019destruction}, we take ResNet-50 and ResNet-101 as the backbones. The results are shown in Table \ref{table:accuracy on fine-grained}, which illustrates that the performance of PAA is consistently better than other methods and PAA has achieved remarkable performance on these challenging fine-grained tasks.

\begin{table*}[t]
\vspace{-1.0em}
\caption{{Comparison of computational cost} (GPU hours) between our proposed PAA and other previous automated DA methods. We train Wide-ResNet-28-10 on CIFAR-10 and ResNet-50 on ImageNet. \emph{Search}: the time of searching for augmentation policys. \emph{Train}: the time of training the target network. 
\emph{Total}: the total time. Except PAA, all metrics are cited from \citep{zhang2019adversarial,lim2019fast}. The computational cost of PAA is estimated on GeForce GTX 1080 Tis while AA, AdvAA are on NVIDIA Tesla P100 and FastAA is on NVIDIA Tesla V100. }
\begin{center}
\setlength{\tabcolsep}{1.4mm}{
\scalebox{0.93}{
\begin{tabular}{c|c|c c c| c}

\hline
\hline
Dataset & GPU hours & AA  & FastAA  & AdvAA  & PAA \\
\hline
\multirow{3}{*}{CIFAR-10} &  Search  & 5000 & 3.5 & \textbf{$\sim$0} & \textbf{$\sim$0} \\

  & Train  & \textbf{6} & \textbf{6} & - & 7.5 \\
  & Total  & 5006 & 9.5 & - & \textbf{7.5} \\
\hline
 \multirow{3}{*}{ImageNet}& Search & 15000 & 450 & \textbf{$\sim$0} & \textbf{$\sim$0} \\
  & Train & \textbf{160} & \textbf{160} & 1280 & 270 \\
  & Total  & 15160 & 610 & 1280 & \textbf{270} \\
\hline
\hline
\end{tabular}}}
\vspace{-1.0em}
\end{center}

\label{table:computing cost}
\end{table*}

\begin{table}[t]
\caption{Test accuracy (\%) on various fine-grained classification datasets including CUB-200-2011 (CUB) \citep{wah2011caltech}, Stanford Cars (Cars) \citep{krause20133d} and FGVC-Aircraft (Aircraft) \citep{maji2013fine}.}
\label{table:accuracy on fine-grained}
\begin{center}

\scalebox{0.93}{
\begin{tabular}{c|c|ccccc|c}
\hline
\hline
Dataset  & Model      & Baseline & Mixup & CutMix & AA &FastAA   & PAA            \\ \hline
 \multirow{2}{*}{CUB}  & ResNet-50  & 85.5 & 86.2   & 86.1 & 86.8 & 86.5 & \textbf{87.5\scriptsize$\pm 0.2$} \\
    & ResNet-101 & 85.6    & 87.7 & 87.9 & 88.1 & 87.9 & \textbf{88.3\scriptsize$\pm 0.2$}  \\ \hline
\multirow{2}{*}{Cars} & ResNet-50  & 93.0    & 93.9 & 94.1 & 94.2& 94.0 & \textbf{94.3\scriptsize$\pm 0.1$} \\
     & ResNet-101 & 93.1    & 94.1 & 94.2 & 94.2 & 93.8 & \textbf{94.5\scriptsize$\pm 0.1$} \\ \hline
\multirow{2}{*}{Aircraft} & ResNet-50  & 91.0    & 92.0 & 92.2 & 92.3 & 92.2 &  \textbf{92.6\scriptsize$\pm 0.2$} \\
     & ResNet-101 & 91.6    & 92.9 & 92.3 & 92.8 & 92.9 & \textbf{93.5\scriptsize$\pm 0.3$} \\ \hline
\hline
\end{tabular}}
\end{center}

\vspace{-1.0em}

\end{table}

\subsection{Complexity Analysis}

In this section, in order to further demonstrate the performance of PAA in terms of complexity, we compare the policy search time and training time of PAA with AA \citep{cubuk2018autoaugment}, FastAA \citep{lim2019fast} and AdvAA \citep{zhang2019adversarial}, as illustrated in Table \ref{table:computing cost}. As shown in Table \ref{table:computing cost}, compared to the previous works, PAA 
requires the fewest total computational resources, and the search time is almost negligible. As for the parameters, the total number of PAA model parameters (about 0.23M) is less than 1$\%$ of the target network (\egno, ResNet50: about 25.5M).

In summary, the main reasons to reduce the computational cost lie in three aspects: 1) The augmentation policy network is jointly optimized with the target network, similar to \citep{zhang2019adversarial,lin2019online}. Namely, our proposed method searches for policies in an online manner, obviating thousands of policies validation on a small proxy network and the requirement for retraining the target network. Besides, we use fixed schedules for the two corresponding hyperparameters (\ieno, probability and magnitude) of each transformation to effectively reduce the search space, which makes it easier to search for effective policies. By these means, PAA compresses most of \textbf{the policy search time} 2) As mentioned before, a MARL algorithm is adopted, in which all agents parallelly learn the augmentation policies, to reduce \textbf{the training time}. 
3) Compared with the previous DA methods using image-by-image sequential transformations, we perform parallel transformations on tensor. Specifically, we pick the patches in a batch performing the same operation to reconstruct a new tensor. And we use Kornia\footnote{Kornia \citep{riba2020kornia} is a differentiable computer vision library for PyTorch. We use it to accelerate augmentation operation on tensors.} to realize tensor transformations on GPU to further reduce \textbf{the processing time} which is included in the training time.

\subsection{Ablation Study}

In this section, we study the effectiveness of MARL and also discuss the design of patch numbers, through several ablation experiments.
\label{subsection:Ablation Study}

\begin{table*}[]
\vspace{-1.0em}
\begin{center}
\caption{Ablation study: performance comparisons ($\%$) of \textit{Patch RandomAugment} (PRA), \textit{Patch Single AutoAugment} (PSAA) and our PAA on CIFAR-10 and CIFAR-100 (test accuracy) with Wide-ResNet-28-10 (WRN-28-10), and on ImageNet (Top-1 / Top-5 accuracy) with ResNet-50.}

\label{table:ablation study}
\scalebox{0.93}{
  \begin{tabular}{c|ccc|ccc}
    \hline
    \hline
    \multirow{2}{*}{Method} & \multicolumn{3}{c|}{Policy}  & \multicolumn{3}{c}{Dataset} \\
\cline{2-7}     & \multicolumn{1}{c}{random} & SARL & MARL  & CIFAR-10 & CIFAR-100 & ImageNet \\
    \hline
    PRA & \checkmark  &   &         & 97.1 & 83.0 & 77.9 / 93.9 \\
    PSAA &  &  \checkmark     &       & 97.3 & 83.1 & 78.0 / 93.9 \\ 
    PAA(Ours) & &   & \checkmark    & \textbf{97.5} & \textbf{83.4} & \textbf{78.3 / 94.1} \\
    \hline
    \hline
\end{tabular}}
\vspace{-0.5em}
\end{center}

\end{table*}

\begin{table}[]

\begin{center}
\caption{Ablation study: performance comparisons (Top-1 accuracy ($\%$)) under five different patch numbers $N$ on CIFAR-10 (image size: $32\times{32}$) with WRN-28-10 and on ImageNet (image size: $224\times{224}$) together with CUB-200-2011 (CUB) (image size: $448\times{448}$) with ResNet-50 backbone.
}
\label{table:different N accuracy}
\scalebox{0.93}{
\begin{tabular}{c|ccccc}
\hline\hline
N            & \multicolumn{1}{c}{4} & \multicolumn{1}{c}{16} & \multicolumn{1}{c}{64} & \multicolumn{1}{c}{256} & \multicolumn{1}{c}{1024} \\ 
\hline
CIFAR-10 & \textbf{97.5}                & 97.3               & 97.2               & 97.1                 & -                  \\ \hline
CUB & 87.2                & \textbf{87.5}               & 87.3               & 87.1                 & 86.8                  \\ \hline
ImageNet     &    78.1              & \textbf{78.3}                 & 78.2                 & 78.0                 & 78.0                  \\
\hline\hline
\end{tabular}}
\vspace{-1.0em}
\end{center}

\end{table}

\begin{table}[t]
\vspace{-2.0em}
\begin{tabular}{cc}
\includegraphics[width=2.4in]{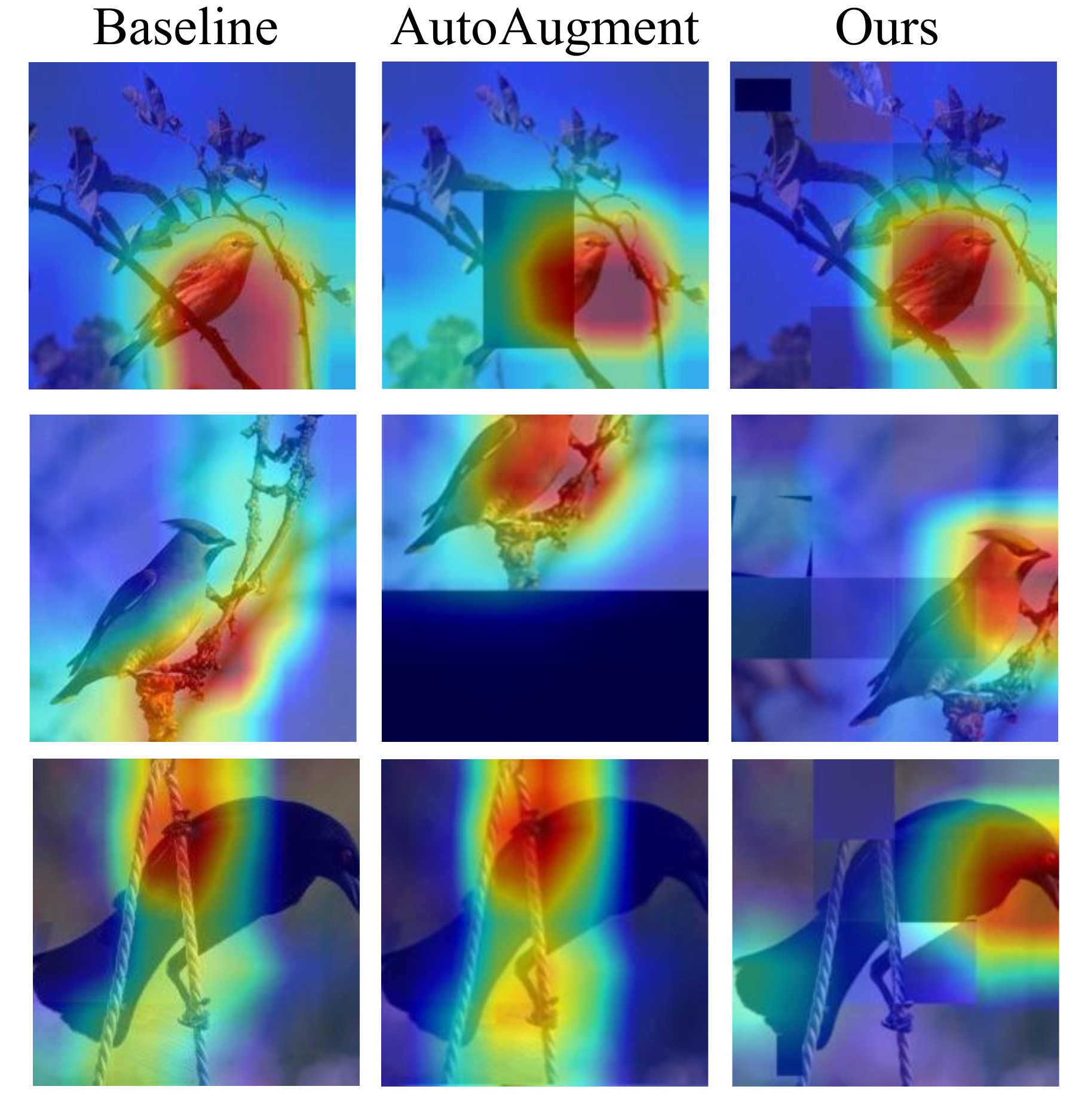}  &  \includegraphics[width=2.8in]{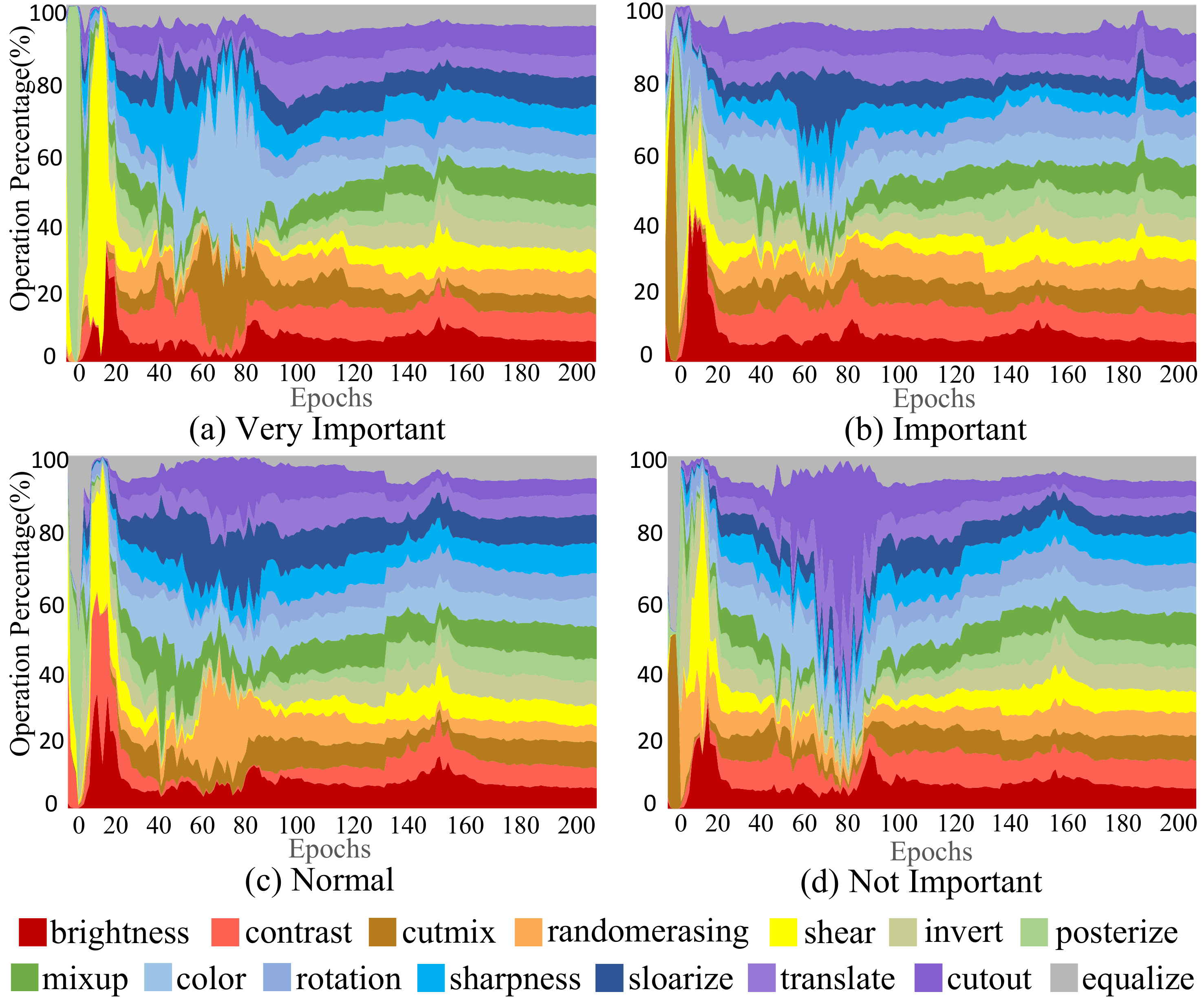}
\end{tabular} 
\end{table}

\begin{figure}[t]
\begin{minipage}[t]{0.46\linewidth} 
\vspace{-1.4em}
\centering 
\captionsetup{width=0.95\linewidth}
\caption{Grad-CAM visualization on the examples from CUB with ResNet-50 trained by baseline, AutoAugment and our PAA. PAA tends to help the target network focus on task-related features.} 
\label{fig:gradcam} 
\end{minipage}%
\begin{minipage}[t]{0.56\linewidth} 
\centering 
\vspace{-1.4em}
\captionsetup{width=0.93\linewidth}
\caption{We categorize patches into four groups according to the patch importance calculated through Grad-CAM.
We draw four \textbf{stacked area chart of the operations' percentages} over time. The area represents the percentage of each operation.} 
\label{fig:policy} 
\end{minipage}
\vspace{-2.0em}
\end{figure}

\textbf{Performance of random policies.} We randomize the augmentation policy for each patch, dubbed \textit{Patch RandAugment} and compare it with our proposed PAA. To be specific, the augmentation policy is randomly selected from predefined transformations with uniform probability. As shown in Table \ref{table:ablation study}, \textit{Patch RandAugment} (PRA) leads to considerable appreciable improvements. However, our Patch AutoAugment with meaningful guidance significantly surpasses the random patch policies.

\textbf{Performance of the policies searched using single-agent RL.} Furthermore, we directly use the same single-agent RL to search for each patch's policy, termed as \textit{Patch SingleAutoAugment} (PSAA) and compare it to our PAA, where the search task is a cooperative multi-agent problem. In short, PSAA ignores the contextual relationship between patches, where patches are non-cooperative. The results (see Table~\ref{table:ablation study}) indicate that taking the contextual information into account with cooperative RL-agents has improved the joint effectiveness of DA.

\textbf{Impact of the number of patches.}
We explore the effect of the number of patches on the target model performance. As shown in Table \ref{table:different N accuracy}, we respectively set the number of patches $N=\{4,16,64,256,1024\}$. The results show that on ImageNet / CUB-200-2011, when patch numbers $N$ increase, the performance accuracy first increases and then decreases. In addition, when $N=16$, PAA achieves the best performance. 
We analyze that the small number of patches may cause PAA to be unable to effectively explore local diversity, and its advantages would be limited. In particular, when $N=1$, PAA degenerates into image-wise automated DA. In contrast, under too larger values (\egno, the extreme case is to search for DA policy for each pixel), the local semantic consistency is broken and the benefit brought by the consideration of contextual relationship between patches is gradually overtaken.

\subsection{Visualization}\label{subsection:visualization}

\textbf{Policy Visualization.}
Grad-CAM \citep{selvaraju2017grad} is used to localize the important regions in the image. Therefore we can calculate the importance score of each patch, and we divide the importance scores into four bins, \ieno, \textit{very important, important, normal} and \textit{not important}. Then, we categorize patches into four groups and draw four stacked area charts showing the percentages of operations selected by PAA augmentation policies over time. We take ResNet-50 backbone trained on CUB-200-2011 as an example. 

Since our proposed PAA searches for patch policies in an online manner, the strategies change dynamically over time as shown in Figure \ref{fig:policy}. At the beginning of the training process, the selected actions are messy since the MARL network is in the exploratory stage. At the tail end of the training, the target network has converged, causing the percentage of all operations to be almost the same and the percentage to flatten out.

In addition, we have some interesting findings which may provide some insights to the DA community. $i$) The optimal augmentation strategies of patches vary by their important levels. Therefore, it is necessary to take the image content into account when performing data augmentation. $ii$) In the middle of the training process, different types of patches prefer to select different augmentation operations. Concretely, as illustrated in Figure \ref{fig:policy}, for the important patches, \textit{color} transformation is mostly picked. For the unimportant patches, \textit{RandomErasing} and \textit{Cutout} are usually chosen by PAA. Important patches commonly take along semantic information that is related to mainstream tasks. It is better to choose the mild transformations (\egno, \textit{color}) for them, which can effectively protect the semantic information from being damaged. In contrast, unimportant patches typically carry unexpected features \citep{singla2021understanding} which are causally unrelated to the desired class. Severe transformations (\egno, \textit{Cutout}, RandomErasing) could be chosen for them, which introduce noise and disturbance to reduce the impact of unexpected features on the target network learning.

\textbf{Grad-CAM Visualization.} Here, we adopt Grad-CAM \citep{selvaraju2017grad} to visualize the learned features to intuitively show the impact of PAA, as shown in Figure \ref{fig:gradcam}. We take the ResNet-50 as the backbone which is trained with dataset processed by 1) Baseline 2) AutoAugment and 3) Patch AutoAugment, respectively. We observe that the model trained with PAA focus on more task-related areas rather than spurious correlations (\egno, the branch where the bird stands) or overemphasized features (\egno, birds' claws). More visualization results are provided in the Appendix \ref{appendix:gradcam}.

\section{Conclusion}

In this paper, we propose Patch AutoAugment (PAA), a more fine-grained automated data augmentation approach. Our method adopts multi-agent reinforcement learning to automatically search for the optimal augmentation policies for patches, and encourages agents to cooperate with each other to further achieve the joint optimal policy across the entire image. Extensive experiments demonstrate that PAA improves the target network performance with low computational cost in various tasks. Meanwhile, we use visualization to show that PAA is beneficial for the target network to localize more class-related cues.  Furthermore, we hope that our visual observations of policies will be useful for future development.
In future work, we will investigate different schemes on dividing different regions.
Furthermore, our method is naturally aligned with the patch token mechanism in the current vision transformers \citep{dosovitskiy2020image,touvron2021training,liu2021swin} and the data augmentation specific to vision transformers has not been extensively studied. Therefore, we leave the automated data augmentation for vision transformers to future work, which may provide some interesting insights to the community.

\bibliography{iclr2022_conference}
\bibliographystyle{iclr2022_conference}

\newpage
\section*{Appendix}
\appendix

\begin{table}[]
\vspace{-0.5em}
\caption{We list fifteen kinds of augmentation operations that we use. Additionally, the schedule of magnitude (\ieno, the parameters for Kornia \citep{riba2020kornia} pytorch library) for each operation are shown in the third column. Some transformations do not use the magnitude information (\egno, CutMix and Cutout).}
\label{table:operations}
\begin{center}
\setlength{\tabcolsep}{1.5mm}{
\scalebox{0.85}{
\begin{tabular}{c|l|c}
\hline
Operation Name & \multicolumn{1}{c|}{Description}                                                                                                                                          & magnitudes                                                            \\ \hline
Brightness     & \begin{tabular}[c]{@{}l@{}}Adjust the brightness of the patch. A magnitude=0 gives a\\ black patch, whereas magnitude=1 gives the original patch.\end{tabular}            & brightness=(0.5, 0.95)                                                            \\ \hline
Contrast       & \begin{tabular}[c]{@{}l@{}}Control the contrast of the patch. A magnitude=0 gives a gray\\ patch, whereas magnitude=1 gives the original patch.\end{tabular}              & contrast=(0.5, 0.95)                                                              \\ \hline
CutMix         & \begin{tabular}[c]{@{}l@{}}Replace this patch with another patch (selected at random \\ from the patches which are also performed \textit{CutMix}).\end{tabular}                                            & -                                                                                 \\ \hline
Cutout      & Set all pixels in this patch to the average value of the patch.                                                                                                           & -                                                                                 \\ \hline
Invert         & Invert the pixels of the patch                                                                                                                                            & -                                                                                 \\ \hline
Mixup        & \begin{tabular}[c]{@{}l@{}}Linearly add the image with another image (selected at random \\ from the patches which are also performed \textit{Mixup}). \end{tabular} & $\lambda \sim Beta(1,1)$                                                                               \\ \hline
Posterize      & Reduce the number of bits for each pixel to magnitude bits.                                                                                                               & bits=3                                                                            \\ \hline
Solarize      & Invert all pixels above a threshold value of magnitude.                                                                                                               & thresholds=0.1                                                                        \\ \hline
RandomErasing  & Erases a random rectangle region in a patch.                                                                                                                                     & \begin{tabular}[c]{@{}c@{}}scale=(0.09, 0.36),\\  ratio=(0.5, 1/0.5)\end{tabular} \\ \hline
Rotation       & Rotate the patch magnitude degrees.                                                                                                                                       & degrees=30.0                                                                      \\ \hline
Sharpness      & \begin{tabular}[c]{@{}l@{}}Adjust the sharpness of the image. A magnitude=0 gives a\\ blurred image, whereas magnitude=1 gives the original image.\end{tabular}           & sharpness=0.5                                                                     \\ \hline
Shear(X/Y)        & \begin{tabular}[c]{@{}l@{}}Shear the image along the horizontal or vertical axis with rate \\ magnitude\end{tabular}                                                      & shear=(-30, 30)                                                                   \\ \hline
Translate(X/Y)      & \begin{tabular}[c]{@{}l@{}}Translate the patch in the horizontal or vertical direction \\ by absolute fraction of patch length.\end{tabular}                              & translate=(0.4, 0.4)                                                              \\ \hline
Color          & Adjust the color balance of the image.                                                                                                                                    & hue=(-0.3, 0.3)                                                                   \\ \hline
Equalize       & Equalize the image histogram.                                                                                                                                             & -                                                                                 \\ \hline
\hline
\end{tabular}}}
\end{center}  

\end{table}

\begin{table}[]
\caption{The model architecture of actor network and critic network in PAA. For each convolution layer, we list the input dimension, output dimension, kernel size, stride, and padding. For the fully-connected layer, we provide the input and output dimension. BN is short for batch normalization. }
\label{table:architecture}
\begin{center}
\setlength{\tabcolsep}{0.9mm}{
\scalebox{0.85}{
\begin{tabular}{ccc}
\hline
\multicolumn{1}{c|}{Layer} & \multicolumn{1}{c|}{Actor network}                    & Critic network \\ \hline
\multicolumn{1}{c|}{1}     & \multicolumn{1}{c|}{ReLU,Conv2D(32,64,3,1,1),BN}      & FC(1568,256)   \\ \hline
\multicolumn{1}{c|}{2}     & \multicolumn{1}{c|}{ReLU,Conv2D(64,64,3,1,1),BN}      & ReLU           \\ \hline
\multicolumn{1}{c|}{3}     & \multicolumn{1}{c|}{ReLU,Conv2D(64,15,3,2,1),Softmax} & FC(256,1)      \\ \hline
\end{tabular}}}
\end{center}
\vspace{-0.5em}
\end{table}

\section{Operations Details}
\label{appendix:operation}
Following \citep{cubuk2018autoaugment,ho2019population,lim2019fast,zhang2019adversarial}, we define the fifteen common augmentation operations to form the action space. Here, we detail the description of these operations, as illustrated in Table \ref{table:operations}. In addition, we give magnitudes range of augmentation operations corresponding to hyperparameters of the functions in the Kornia PyTorch library. Some operations, such as Cutout and CutMix, have no parameters.

Additionally, in our implementation, we pick out the patches that perform the same operation and put them into a new tensor. Then, we speed up the process by performing parallel transformations on the tensor. Tensor transformations on GPU can be realized by Kornia \citep{riba2020kornia} to reduce the computational costs. In particular, when performing the label-disturbing operations (\ieno, Mixup and CutMix), a patch needs to mix with another patch that is randomly selected from the new tensor. Namely, a patch is mixed with another patch that performs the same operation.

\begin{table*}[]
\vspace{-0.5em}
\caption{The hyperparameters of various target models on CIFAR-10, CIFAR-100, ImageNet, CUB-200-2011, Stanford Cars and FGVC-Aircraft. LR represents learning rate of the target network, WD represents weight decay, and LD represents learning rate decay method. If LD is multistep, we decay the learning rate by 10-fold at epochs 30, 60, 90 etc. according to LR-step. LR-A2C represents the learning rate of augmentation model. }
\label{table:hyperparameters}
\begin{center}
\scalebox{0.85}{
\setlength{\tabcolsep}{1.8mm}{\begin{tabular}{c|c|c c c c c c|c}
\hline
\hline
Dataset&Model & BatchSize & LR & WD & LD &LRstep&LR-A2C& Epoch \\
\hline
\multirow{5}{*}{CIFAR-10} & WRN-28-10 & 128 & 0.1 & 5e-4 & cosine &-&1e-3 &200  \\
&SS(26 2x32d) & 128 & 0.2 & 1e-4 & cosine&-&1e-4 &600\\
&SS(26 2x96d) &128 & 0.2 &1e-4 & cosine &-&1e-4 &600\\
&SS(26 2x112d) &128 & 0.2 & 1e-4 &cosine&- &1e-4 &600\\
&Pyramid+SD & 128 & 0.1 & 1e-4 &cosine&- &1e-4 &600 \\
\hline
\multirow{3}{*}{CIFAR-100} &WRN-28-10&128&0.1&5e-4&cosine&-&  1e-4&200\\
&SS(26 2x96d) &128 &0.1 &5e-4&cosine &-&1e-4&1200\\
&Pyramid+SD & 128 &0.5 &1e-4 &cosine&-&1e-4&  1200 \\
\hline
\multirow{2}{*}{ImageNet} &ResNet-50&512&0.1&1e-4&multistep&[30,60,90,120,150]& 1e-4&270\\
&ResNet-200 &512 &0.1 &1e-4&multistep& [30,60,90,120,150]&1e-4 &270\\
\hline
\multirow{2}{*}{CUB-200-2011} & ResNet-50&512&1e-3&1e-4&multistep&[30,60,90]& 1e-4&200\\
&ResNet-101 &512 &1e-3 &1e-4&multistep&[30,60,90] &1e-4 &200\\
\hline
\multirow{2}{*}{Stanford Cars} &ResNet-50&512&1e-3&1e-4&multistep&[30,60,90]&1e-4& 200\\
&ResNet-101 &512 &1e-3 &1e-4&multistep&[30,60,90] &1e-4& 200\\
\hline
\multirow{2}{*}{FGVC-Aircraft} & ResNet-50&512&1e-3&1e-4&multistep&[30,60,90]& 1e-4&200\\
&ResNet-101 &512 &1e-3 &1e-4&multistep&[30,60,90] &1e-4 &200\\
\hline
\hline
\end{tabular}}}
\end{center}  

\end{table*}

\section{Datasets}
\label{appendix:dataset}
We evaluate Patch AutoAugment (PAA) on the
following datasets: CIFAR-10 \citep{krizhevsky2009learning},  CIFAR-100 \citep{krizhevsky2009learning}, ImageNet \citep{deng2009imagenet} and three fine-grained image recognition datasets (CUB-200-2011 \citep{wah2011caltech}, Stanford Cars \citep{krause20133d} and FGVC-Aircraft \citep{maji2013fine}).

To be specific, both CIFAR-10 and CIFAR-100 have 50,000 training examples. Each image of size $32 \times 32$ belongs to one of 10 categories. ImageNet dataset has about 1.2 million training images and 50,000 validation images with 1000 classes. Original ImageNet data have different sizes and we resize them to $224 \times 224$. In addition, we evaluate the performance of our proposed PAA on three standard fine-grained object recognition datasets. CUB-200-2011 \citep{wah2011caltech} consists of 6,000 train and 5,800 test bird images distributed in 200 categories. Stanford Cars \citep{krause20133d} contains 16,185 images in 196 classes. The FGVC-Aircraft dataset contains 10,200 images of aircraft, with 100 images for each of 102 different aircraft model variants, most of which are airplanes. The image size in the above three datasets is $448 \times 448$.

\section{Model Architecture}
\label{appendix:model}
Here, we provide the detailed model architecture for each component in our PAA augmentation policy model, including feature extractor network, actor network and critic network. We use the pre-trained on ImageNet ResNet-18 backbone (excluding the final avgpool and softmax layer) to extract the deep features of the image and the patch, which are denoted as the state $s$ and the observation $o_i$ respectively. As for the actor network and the critic network, the detailed model architectures are shown in Table \ref{table:architecture}.

\section{Hyperparameters}
\label{appendix:parameters}
We detail various target models hyperparameters (\egno, batch size, learning rate and training epochs) on CIFAR-10, CIFAR-100, ImageNet, CUB-200-2011, Stanford Cars and FGVC-Aircraft in Table \ref{table:hyperparameters}. We do not specifically tune these hyperparameters, and all of these are consistent with previous works \citep{cubuk2018autoaugment,lim2019fast,dabouei2021supermix,du2020fine,chen2019destruction}. In our PAA model, we set time horizon $T=1$, \ieno, the augmentation policies take actions at every time step for more precise control. In addition, we use the SGD optimizer with an initial learning rate of 1e-4 to train the actor network and the critic network. 

\section{More Grad-CAM Visualization}
\label{appendix:gradcam}
\begin{figure*}[h]

	\begin{center}
		\scalebox{0.93}{
		\includegraphics[width=13cm]{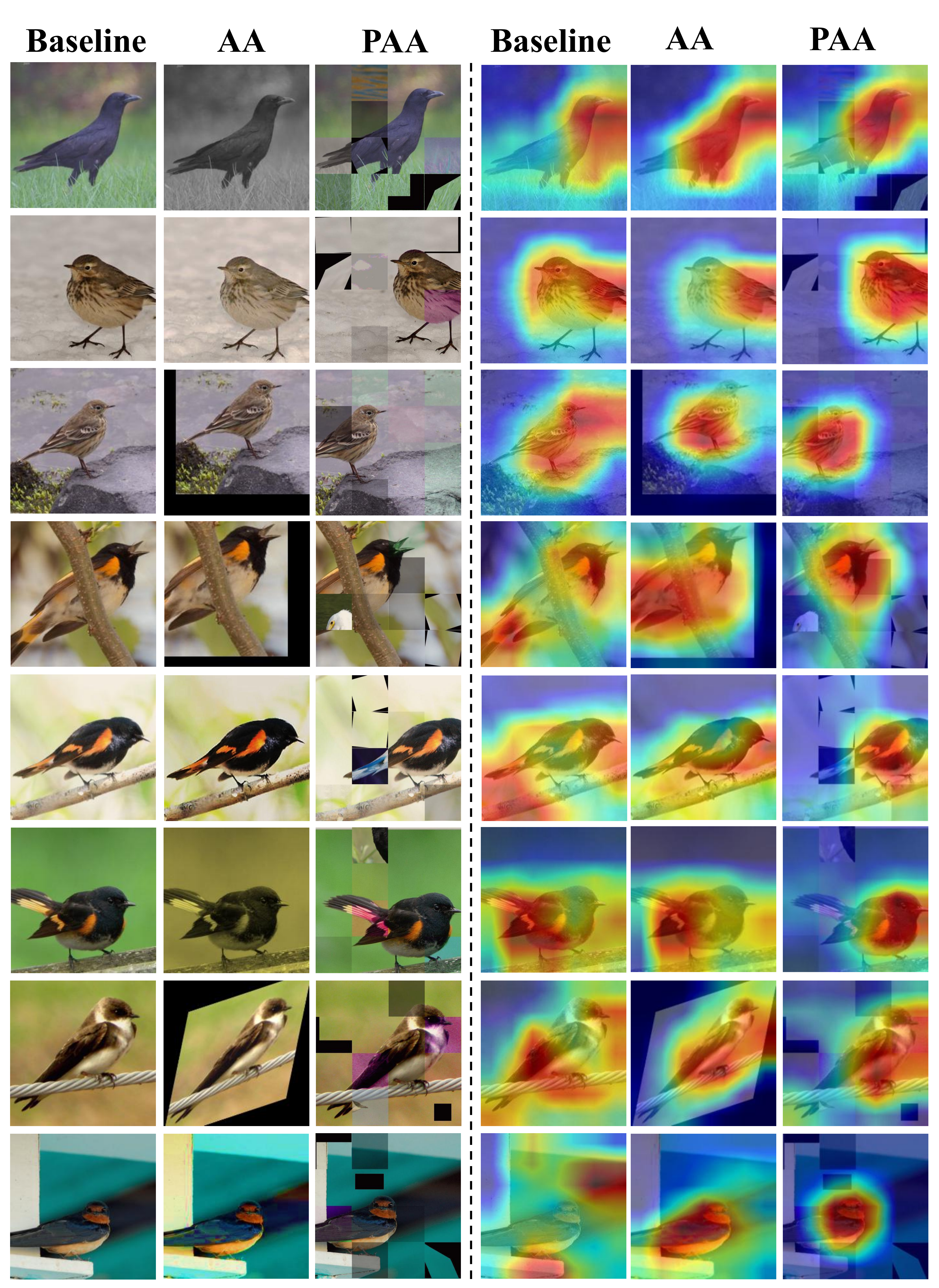}}
	\end{center}
	\caption{ We give more visualizations of augmentation and Grad-CAM\citep{selvaraju2017grad} results of baseline, AA\citep{cubuk2018autoaugment} and our PAA. PAA performs rich diversity of DA and tends to help the target model focus on the class-related features.  }
	\label{fig:baselineaapaa}
\end{figure*}

In this section, we provide more Grad-CAM \citep{selvaraju2017grad} results of ResNet-50 models trained using baseline augmented data, AA \citep{cubuk2018autoaugment} and our proposed PAA, respectively, as shown in Figure \ref{fig:baselineaapaa}.
The visualization results demonstrate that our proposed PAA improves the localization ability of the target network and tends to help the target network focus on more parts of the foreground object. In short, our proposed PAA make the target network focus on the important or representative regions closely related to the class within an image.

\end{document}